\documentclass[twocolumn,10pt]{article}

\usepackage[T1]{fontenc}
\usepackage[utf8]{inputenc}
\usepackage{amsmath,amsfonts,amssymb}
\usepackage{graphicx}
\usepackage{booktabs}
\usepackage{geometry}
\usepackage{cite}
\usepackage{abstract}
\usepackage{multirow}
\usepackage{threeparttable}
\usepackage{placeins}
\usepackage{caption}
\usepackage{titlesec}
\usepackage{hyperref}
\usepackage{stfloats}
\usepackage{flafter}
\usepackage{subcaption}
\titlespacing*{\section}{0pt}{1.0ex plus 0.3ex minus 0.2ex}{0.8ex plus 0.2ex}
\titlespacing*{\subsection}{0pt}{0.8ex plus 0.2ex minus 0.2ex}{0.5ex plus 0.2ex}

\hypersetup{colorlinks=true, linkcolor=blue, citecolor=blue, urlcolor=blue}
\title{\textbf{An Empirical Study of Data Scale, Model Complexity, and Input Modalities in Visual Generalization}}

\author{
    Yidi Zhouluo \\
    School of Medical Information and Artificial Intelligence, \\
    Shandong First Medical University, Taian 271000, Shandong, China \\
    \texttt{zhouluoyidi@gmail.com}
}

\date{June 2026}

\begin{document}

\twocolumn[
\begin{@twocolumnfalse}

\maketitle

\begin{abstract}
Modern deep neural networks usually have large parameter scales and nonlinear hierarchical structures, and they have achieved strong performance in computer vision. However, the source of their generalization performance remains difficult to explain using traditional statistical learning theory. Among the factors that may affect visual generalization, data scale, model complexity, and input modalities are fundamental and controllable variables. This study empirically analyzes how these three factors influence model generalization performance. Specifically, in a preliminary experiment, we construct a one-dimensional nonlinear function and vary the number of training samples and the polynomial degree to observe the effects of data scale and model complexity on model performance. In the main experiments, we compare model performance on CIFAR-10 and CIFAR-100 under different training data scales, model architectures, and input modalities. The experimental results show that increasing the training data scale consistently improves generalization performance, whereas changes in model complexity do not provide stable gains. In addition, removing color information degrades model performance, while explicit prior features such as gradients, edges, and wavelets have inconsistent effects across different model architectures. Overall, this study provides an empirical analysis of the relationships among data scale, model complexity, input modalities, and visual generalization performance. Code and experimental logs are available at: \url{https://github.com/YidiZhouluo/DeepLearning-Empirical-Studies/tree/main/Exp_01}.
\end{abstract}

\vspace{0.5em}
\vspace{1.5em}

\end{@twocolumnfalse}
]

% --- Section 1: Introduction ---
\section{Introduction}

In recent years, deep neural networks have been widely applied in computer vision, natural language processing, and image/video generation~\cite{krizhevsky2012imagenet,vaswani2017attention,rombach2022high}. With the development of model architectures and the growth of computational resources, over-parameterization has become an important characteristic of modern neural networks. However, traditional machine learning theory often suggests that when model complexity exceeds a certain level, increasing model complexity may lead to degraded generalization performance~\cite{geman1992neural,vapnik1971uniform}. In contrast, existing studies have shown that over-parameterized models can still maintain good test performance while fitting the training data sufficiently well~\cite{zhang2017understanding,belkin2019reconciling,nakkiran2020deep}, and may even exhibit so-called emergent abilities~\cite{wei2022emergent}. These phenomena have motivated a reconsideration of the theoretical foundations of deep learning and the search for explanations of modern neural network generalization.

Empirical research forms hypotheses or tests hypotheses through experimental observation, comparison, and induction. To address the long-standing black-box problem in deep learning, empirical studies have been widely used to explain and analyze representation phenomena in neural networks~\cite{olah2017feature,zeiler2014visualizing}. One common approach is to observe model behavior under controlled data distributions. In such a setting, mathematical and statistical experiments are first conducted on low-dimensional synthetic data to obtain interpretable phenomena or hypotheses, which can then be examined in larger-scale experiments. Among the factors that affect model generalization performance, training data scale, model complexity, and input modalities are fundamental and controllable variables. Training data scale determines the number of empirical samples available for learning. Model complexity affects the upper bound of the function complexity that a model can fit and its representation ability. Input modalities determine the amount and type of information available in the input. For example, introducing prior features such as gradients, textures, and edges can provide additional constraints for model learning~\cite{pandey2020incorporating,li2020wavelet}.

To analyze the effects of these factors on model generalization performance, this study designs a series of controlled experiments. Directly analyzing generalization mechanisms in large-scale neural networks is difficult, because the training process is affected by many factors, including model architecture, optimizer, loss function, learning-rate scheduling, and explicit regularization. Therefore, to obtain a more controlled experimental setting, we first construct a one-dimensional nonlinear function in a preliminary experiment, fit it with polynomials of different degrees, vary the training data scale, and observe model performance. In this study, model performance is mainly measured by training loss, test loss, and test accuracy in classification tasks. Although the conclusions drawn from the preliminary experiment cannot be directly used to prove phenomena in deep neural networks, they can serve as a phenomenological reference for subsequent image classification experiments.

Based on the observations from the preliminary experiment, we further design controlled experiments on the CIFAR-10 and CIFAR-100 image classification datasets~\cite{krizhevsky2009learning}. Specifically, we compare the performance of MLP, AlexNet, and ResNet models under different settings from three perspectives: training data scale, model complexity, and input modalities. Through these experiments, this study aims to answer the following questions. First, does reducing the amount of training data lead to a clear decrease in generalization performance for models of different complexity? Second, does removing color information reduce fitting ability and generalization performance, and do different model architectures show different sensitivity to the absence of color information? Third, can introducing explicit prior features such as edges, gradients, and wavelets improve model fitting ability and generalization performance?

% --- Section 2: Related Work ---
\section{Related Work}

\subsection{Deep Neural Networks and Over-parameterization}

The generalization performance of deep neural networks has long been an important topic in machine learning theory. Existing studies have analyzed this problem from the perspectives of the number of parameters, optimization process, data distribution, and regularization~\cite{zhang2017understanding}. Prior work has also pointed out that the parameter scale and training behavior of modern deep networks make it difficult for classical complexity-based theories to directly explain their generalization behavior~\cite{neyshabur2017exploring,zhang2017understanding}. In addition, data scale has a substantial influence on model performance~\cite{sun2017revisiting,hestness2017deep}. Generalization performance is a key criterion for evaluating the practical utility of machine learning models. Although over-parameterized deep learning models often achieve good generalization performance, the factors that affect their generalization remain insufficiently understood.

Over the past decade, researchers have increasingly used controlled experiments to study representation behavior in deep neural networks. Zhang et al.~\cite{zhang2017understanding} showed that deep neural networks can fit training data even with random labels or randomized inputs. This suggests that classical generalization analysis tools, including VC dimension, Rademacher complexity, and algorithmic stability~\cite{vapnik1971uniform,bartlett2002rademacher,bousquet2002stability}, are insufficient to fully explain the generalization behavior of deep networks in practical tasks. This line of work has encouraged researchers to rethink the sources of deep neural network generalization, especially the relationships among model capacity, optimization, and data distribution.

\subsection{Double Descent, Data Scale, and Model Complexity}

Existing studies have observed that the generalization curves of deep neural networks do not always follow the classical bias-variance trade-off~\cite{geman1992neural,belkin2019reconciling}. The phenomenon of double descent has therefore motivated further investigation into generalization performance. Belkin et al.~\cite{belkin2019reconciling} extended the classical bias-variance curve into the over-parameterized regime and showed that the generalization error may decrease again after model complexity passes the interpolation threshold. Nakkiran et al.~\cite{nakkiran2020deep} further observed double descent behavior associated with model scale, data scale, and training time in deep learning tasks, suggesting that model generalization performance is not determined by a single factor. Motivated by these studies, this work examines the joint effects of training data scale and model complexity on generalization performance in image classification tasks.

\subsection{Visual Model Architectures and Model Capacity}

The success of AlexNet on ImageNet demonstrated that deep convolutional networks can learn effective class-discriminative features for large-scale image classification tasks~\cite{krizhevsky2012imagenet}. The introduction of residual connections enabled the successful training of convolutional networks with more than one hundred layers and led to strong performance~\cite{he2016deep}. Related studies have shown that different network architectures contain different inductive biases, which may affect model generalization performance under the same data conditions~\cite{mitchell1980need,goodfellow2016deep}. Therefore, this study compares MLP, AlexNet, and ResNet models to examine the effects of different model architectures on performance.

\subsection{Input Modalities and Color Features}

Existing studies suggest that the role of color information in input data is task-dependent. In some visual tasks, grayscale input can still allow models to maintain good performance~\cite{bhatta2024achromatopsia}. In contrast, Kanan and Cottrell showed that converting color images to grayscale can affect image recognition performance~\cite{kanan2012color}. Studies on the internal structure of deep neural networks have also shown that convolutional neural networks contain color-sensitive neurons~\cite{engilberge2017color}. This indicates that color information often participates in model decision-making, but its importance may vary with the task objective and model architecture. Therefore, comparing RGB color images with grayscale input under models of different complexity can help analyze the relationship between input information loss and model performance.

\subsection{Introducing Prior Features}

Some studies introduce prior features on top of the original dataset, such as image edges, gradients, or additional features generated by wavelet transforms~\cite{pandey2020incorporating,li2020wavelet}. These methods construct additional input information and provide explicit priors to the model. Prior work has reported that such methods can improve model performance in certain settings, but their benefits often depend on the specific task, model architecture, and training setup. Therefore, this study treats the introduction of additional features as a way to increase input modalities and observes its effect on generalization performance under models of different scales.

% --- Section 3: Preliminary Experiment ---
\section{Preliminary Experiment}

\subsection{Experimental Objective and Method}

To obtain a more controlled experimental setting, this study first constructs a one-dimensional nonlinear function fitting experiment. By varying the polynomial degree, we use the polynomial degree as a proxy variable for model complexity, and then examine how model performance changes under different numbers of training samples. It should be noted that this preliminary experiment is not equivalent to the generalization phenomena observed in deep neural networks. Instead, it serves as a simple and controlled setting for phenomenological observation and provides a reference for the subsequent main experiments.

In the preliminary experiment, we define the following nonlinear function:
\begin{equation}
f(x)=\sin(2\pi x)+0.5\cos(5\pi x), \quad x\in[0,1].
\end{equation}

During data generation, input points $x_i$ are first uniformly sampled from the interval $[0,1]$, and noiseless labels are generated by the target function $f(x)$. To simulate perturbations in real observations, Gaussian noise with mean $0$ and standard deviation $\sigma=0.3$ is then added to the labels:
\begin{equation}
y_i=f(x_i)+\epsilon_i,\quad \epsilon_i\sim\mathcal{N}(0,\sigma^2).
\end{equation}

\begin{figure}[htbp]
    \centering
    \includegraphics[width=0.95\linewidth]{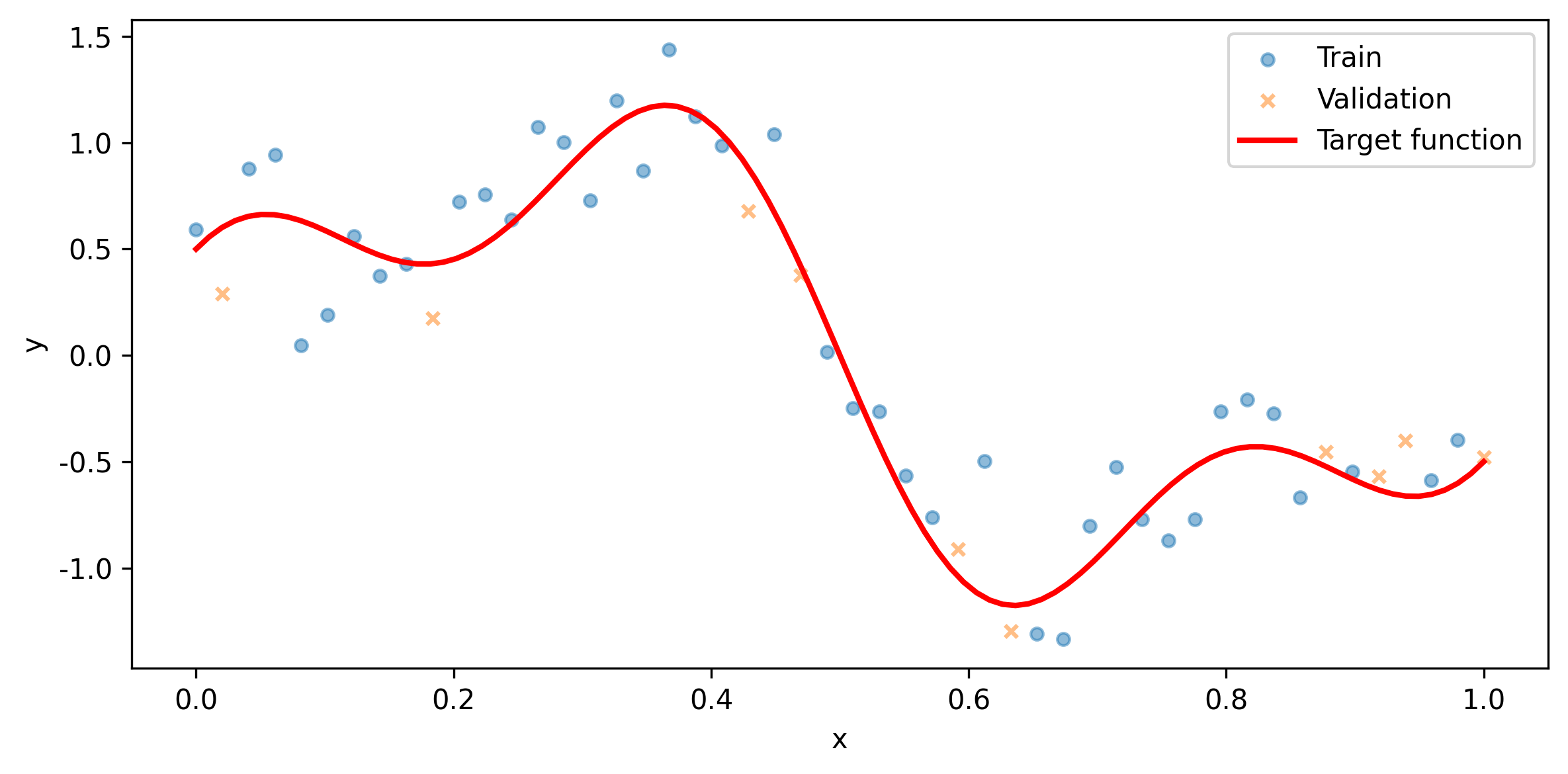}
    \caption{Visualization of the synthetic dataset. Blue dots denote training samples, orange crosses denote validation samples, and the red curve denotes the noiseless target function.}
    \label{fig:1}
\end{figure}

The training and validation sets are split from the noisy samples with a ratio of $8:2$. The test set is generated by uniformly sampling the target function over a fixed interval, with $N_{\text{test}}=100$ samples and without added noise. It is used to evaluate how well the model approximates the true function. The distributions of the training, validation, and test sets are shown in Figure~\ref{fig:1}.

We train polynomial models with degrees from $D=0$ to $D=100$ using gradient descent. For a given degree $D$, the polynomial model is defined as
\begin{equation}
\hat{f}_{D}(x)=\sum_{k=0}^{D}w_kx^k,
\end{equation}
where $D\in\{0,1,\dots,100\}$.

Two data-scale settings are considered. The small-sample setting contains $N_{\text{small}}=50$ training samples, while the larger-sample setting contains $N_{\text{large}}=150$ training samples. To visualize the fitted curves outside the training domain, we also plot the models over the interval $[0,1.1]$. This out-of-domain visualization is used only to support qualitative analysis of in-domain behavior and is not used as a performance evaluation criterion.

During training, the learning rate is set to $\eta=10^{-2}$, the loss function is mean squared error, and gradient descent is run for $30000$ iterations. The mean squared error loss is defined as
\begin{equation}
\mathcal{L}(w)=\frac{1}{n}\sum_{i=1}^{n}\left(\hat{f}_{D}(x_i)-y_i\right)^2.
\end{equation}

The gradient descent update is given by
\begin{equation}
w_{t+1}=w_t-\eta\nabla_w\mathcal{L}(w_t).
\end{equation}

The input $x$ is first mapped to a polynomial feature vector:
\begin{equation}
\phi_D(x)=[1,x,x^2,\dots,x^D].
\end{equation}

To reduce numerical scale differences among polynomial features of different degrees, we standardize the feature matrix. Let $\mu_k$ and $s_k$ denote the mean and standard deviation of the $k$-th feature computed on the training set. The standardized feature is then defined as
\begin{equation}
\tilde{\phi}_{D,k}(x)=\frac{\phi_{D,k}(x)-\mu_k}{s_k}.
\end{equation}
The validation and test sets are transformed using the $\mu_k$ and $s_k$ computed from the training set, which keeps feature scales consistent across data splits and avoids information leakage from validation or test statistics.

\subsection{Experimental Results}

\begin{figure}[!t]
    \centering
    \includegraphics[width=0.95\linewidth]{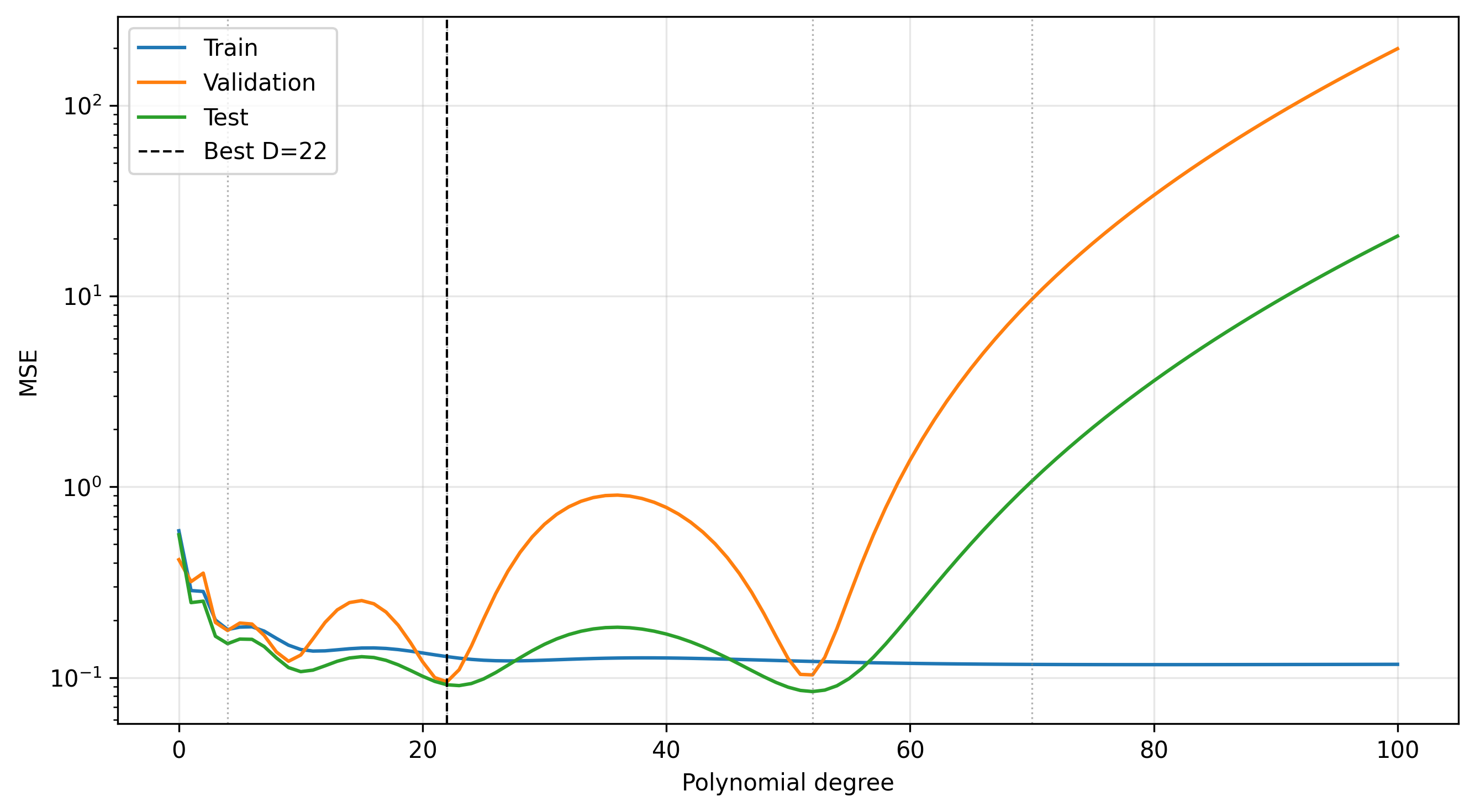}
    \caption{MSE curves over polynomial degrees under the small-sample setting ($N=50$). The dashed vertical line indicates the degree selected by the validation MSE.}
    \label{fig:2}
\end{figure}

Figure~\ref{fig:2} shows the training results under the $N=50$ setting. As the polynomial degree increases, the training loss decreases rapidly at first and then remains nearly unchanged after the model reaches a certain level of complexity. However, the validation and test errors exhibit clear non-monotonic behavior and show a trend similar to double descent. This observation indicates that, under this experimental setting, the relationship between model complexity and test error is not simply monotonic, which differs from the classical bias-variance trade-off. In addition, the best model in Figure~\ref{fig:2} is selected according to the lowest validation error. In fact, on the test set, the lowest error in the second descent region is smaller than the test error of the validation-selected model.

\begin{figure}[!t]
    \centering
    \includegraphics[width=0.95\linewidth]{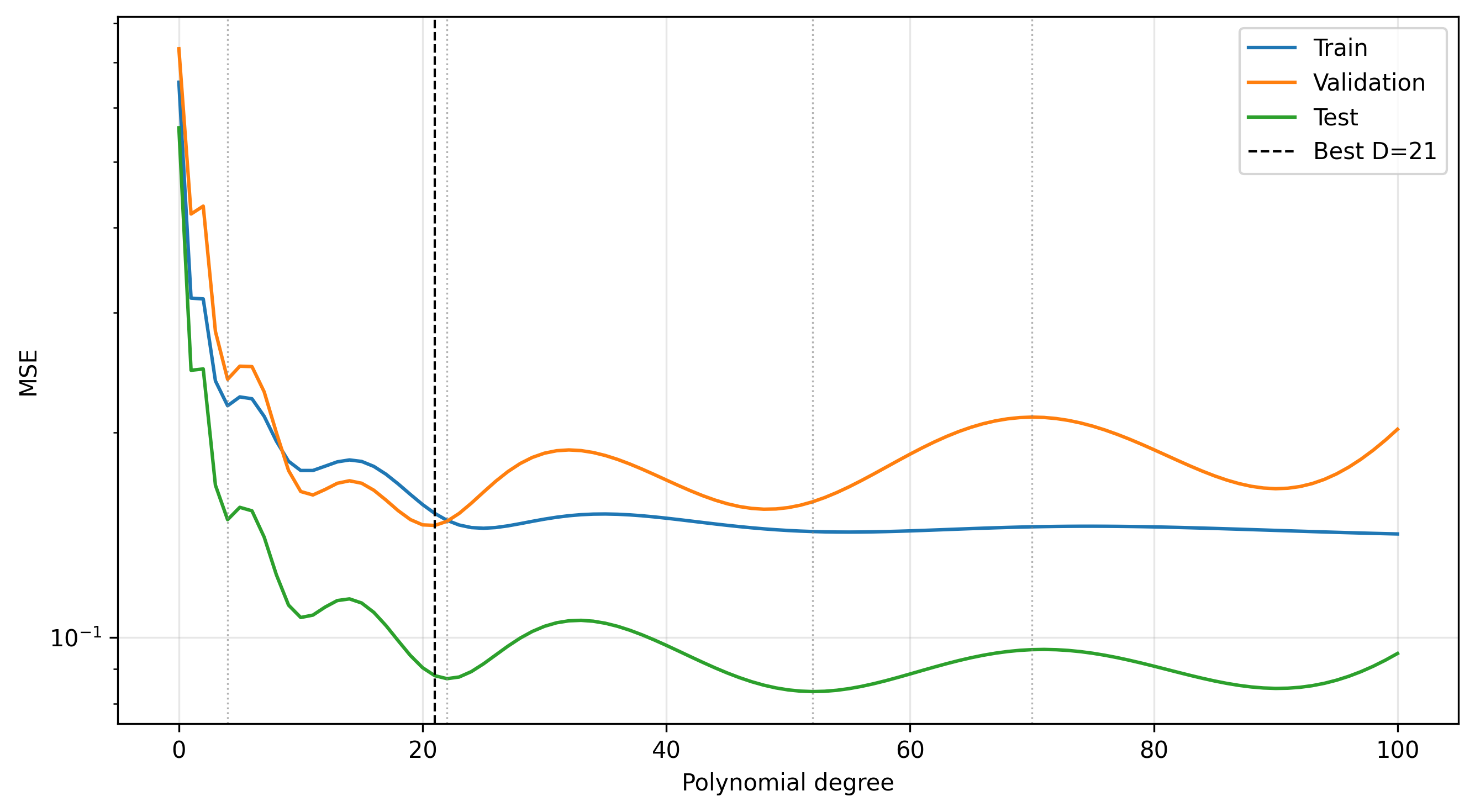}
    \caption{MSE curves over polynomial degrees under the larger-sample setting ($N=150$). The dashed vertical line indicates the degree selected by the validation MSE.}
    \label{fig:3}
\end{figure}

Nevertheless, the double descent-like region observed in the small-sample setting is relatively narrow. In contrast, as shown in Figure~\ref{fig:3}, the test error curve under the $N=150$ setting does not show the same high-degree error explosion observed in the small-sample setting. Instead, the test error remains relatively low across the range of model complexity and exhibits mild alternating decreases and increases as the polynomial degree grows. Notably, in the larger-sample setting, the local minimum test error of high-degree polynomials is still lower than that of low-degree polynomials.

\begin{figure}[!t]
    \centering
    \includegraphics[width=0.95\linewidth]{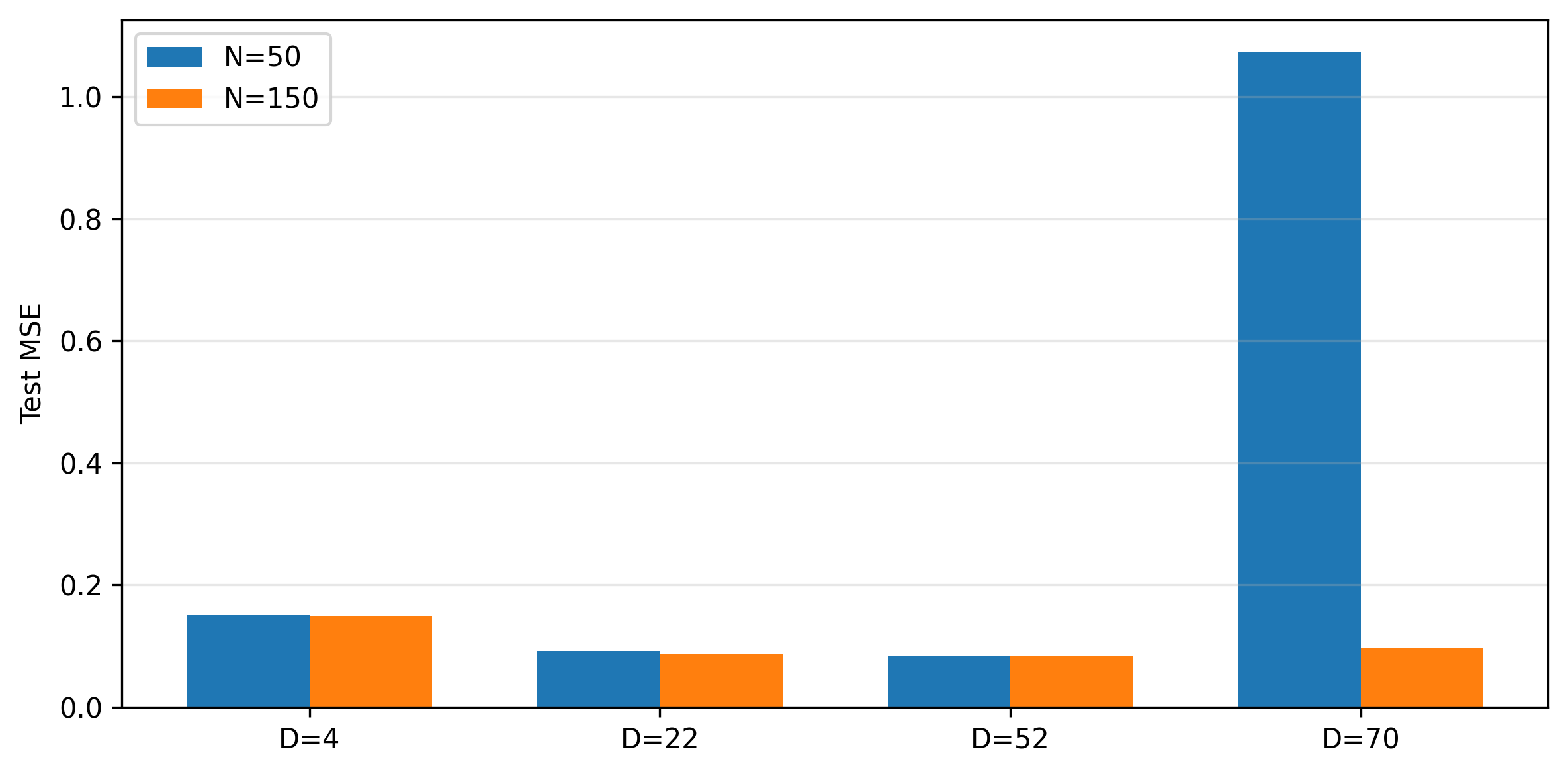}
    \caption{Test MSE of representative polynomial degrees under different training set sizes.}
    \label{fig:4}
\end{figure}

Figure~\ref{fig:4} compares the test errors of four representative polynomial degrees. At $D=4$, $D=22$, and $D=52$, the difference in test error between the two training data scales is relatively small, indicating that increasing the number of training samples does not substantially reduce the test error at these degrees. In contrast, at $D=70$, the test error of the $N=50$ setting increases sharply, while the $N=150$ setting still maintains a relatively low test error.

\begin{figure*}[!t]
    \centering
    \includegraphics[width=0.95\linewidth]{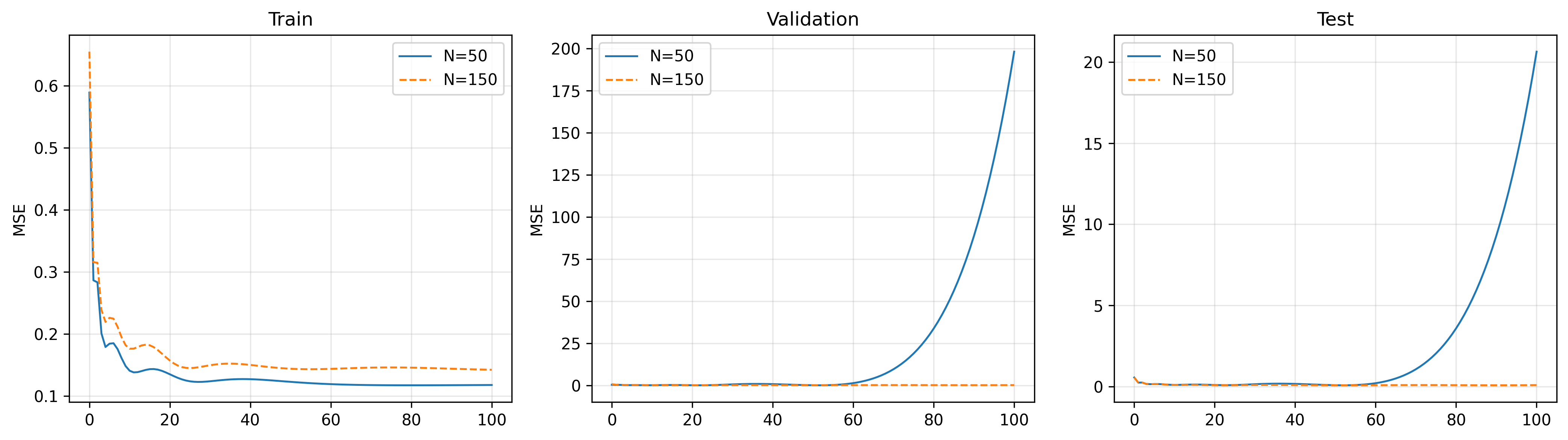}
    \caption{Comparison of train, validation, and test MSE curves between $N=50$ and $N=150$ across polynomial degrees.}
    \label{fig:5}
\end{figure*}

Figure~\ref{fig:5} shows the training, validation, and test performance of polynomial models with different degrees under the two training data scales. Due to the effect of random noise, models trained under the larger-sample setting have higher training error than those trained under the small-sample setting. However, as the polynomial degree increases, the validation and test errors in the small-sample setting rise rapidly after approximately $D\approx60$.

\subsection{Analysis and Discussion}

\begin{figure*}[t]
    \centering
    \includegraphics[width=0.95\linewidth]{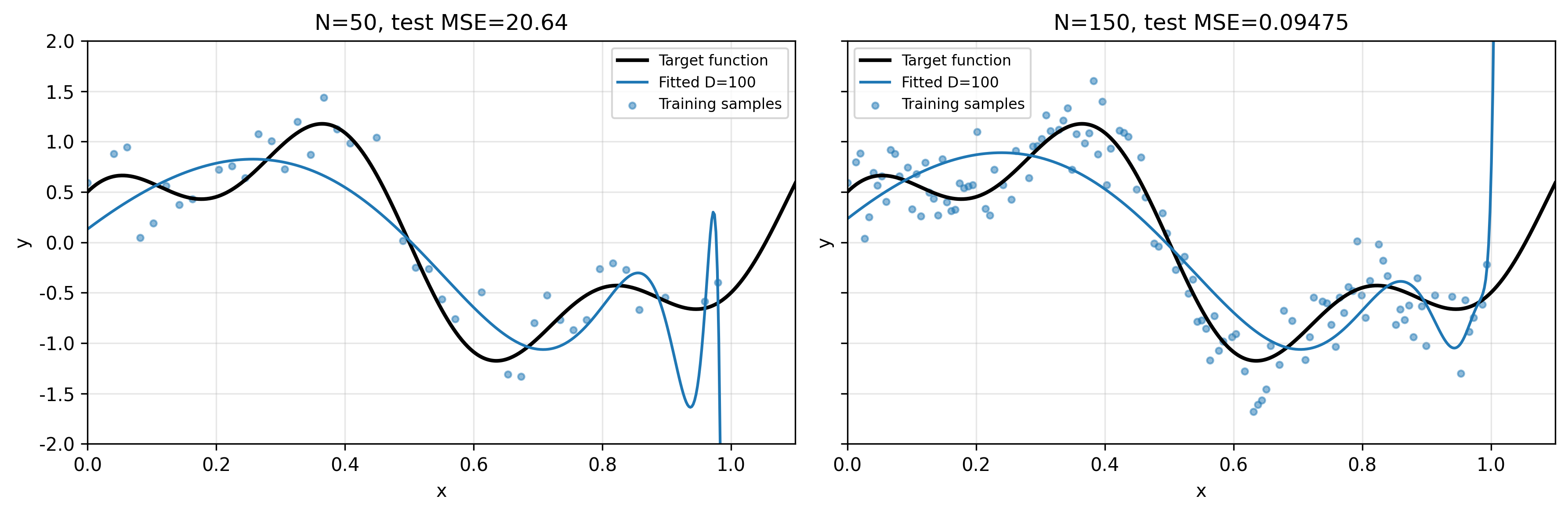}
    \caption{Fitted curves of the $D=100$ polynomial model under different training set sizes. The curves are plotted on $[0,1.1]$ to visualize slight out-of-domain behavior.}
    \label{fig:6}
\end{figure*}

Figure~\ref{fig:6} shows the fitted curves of the $D=100$ polynomial model under the two training data scales. The small-sample setting exhibits strong oscillations near the right boundary of the interval and in the slight out-of-domain region, with a test MSE of 20.64. In comparison, the fitted curve under the $N=150$ setting is closer to the target function overall, with a test MSE of 0.09475. From Figure~\ref{fig:6}, the large test error of the high-degree polynomial may be related to insufficient constraints from training samples near the right boundary. Within the in-domain regions between adjacent training samples, the fitted curve in the larger-sample setting also shows smaller oscillations than that in the small-sample setting.

The above results show that, in this polynomial fitting task, the number of training samples has an important influence on the stability of high-complexity models. When the number of training samples is small, high-degree polynomials can fit the training samples well, but their behavior in sample gaps and near boundaries is insufficiently constrained. As a result, they are more likely to exhibit local oscillations and increased test error. In contrast, a larger training data scale constrains the fitted curve over a denser input interval, allowing high-degree models to maintain more stable test performance within a certain range.

The non-monotonic behavior of validation and test errors indicates that increasing model complexity does not necessarily lead to a monotonic increase or decrease in test error. This phenomenon is similar to the non-monotonic generalization curves studied in double descent, but this preliminary experiment should not be interpreted as direct evidence for the double descent mechanism in deep networks. More precisely, it only shows that, in a low-dimensional and controlled function fitting task, data scale and model complexity can have observable interacting effects.

Therefore, the preliminary experiment provides two motivations for the subsequent image classification experiments. First, when comparing models of different scales, training data scale should be considered jointly rather than discussing model capacity in isolation. Second, changes in the number of training samples may also change the constraints available to the model, thereby affecting its generalization behavior under different levels of model complexity. Based on these observations, the main experiments further construct cross-comparisons between data scale and model complexity to analyze their joint effects on generalization performance. In addition, input modality controlled experiments are designed to examine how reducing original input information and introducing explicit prior features affect model performance.

% --- Section 4: Main Experimental Method ---
\section{Main Experimental Method}

\subsection{Datasets}

To further examine the phenomena observed in the preliminary experiment on real visual tasks, this study conducts experiments on the CIFAR-10 and CIFAR-100 datasets. Both datasets consist of $32\times32$ color images and contain 50000 training images and 10000 test images. CIFAR-10 contains 10 target classes, whereas CIFAR-100 contains 100 target classes. Therefore, compared with CIFAR-10, CIFAR-100 has more categories and fewer available training samples per class, making it a more difficult image classification task. These two datasets allow us to observe model performance under different levels of task difficulty.

\subsection{Controlled Experiments on Data Scale and Model Complexity}

To analyze the relationship between training data scale and model complexity, this study constructs a two-dimensional experimental matrix consisting of training data scale and model architecture. For training data scale, we define five training-set sizes from the 50000 official training images:
\[
N\in\{5000,10000,20000,30000,50000\}.
\]
To reduce additional variation caused by random sampling across different training subsets, we use a fixed random seed to generate a single random permutation of the official training-set indices. For each training data scale, the first $N$ samples from this permutation are selected. This strategy ensures that smaller training subsets are contained within larger training subsets.

For model architecture, three types of models are selected for comparison: an MLP with three hidden layers, an AlexNet adapted for image classification, and a series of ResNet models. The number of parameters and computational complexity of these models are reported in the experimental results section. All models are adapted for CIFAR-style datasets. For example, downsampling operations are reduced, and large-kernel convolutions are replaced with $3\times3$ convolutions. It should be noted that this study assumes that model generalization performance is not determined solely by the number of parameters. Apart from external factors such as explicit regularization, architectural components such as convolutional layers, residual connections, and normalization modules may also introduce different inductive biases. Therefore, data augmentation, weight decay, Dropout, and other explicit regularization strategies are not used in these experiments, in order to reduce the influence of external factors on model performance.

\subsection{Controlled Experiments on Input Modalities}

To investigate the relationship between model performance and input modalities, this study designs controlled input-modality settings. In the color information reduction experiment, color images are converted to grayscale using Eq.~\eqref{eq:rgb2gray}. To keep the input channels consistent with the model's first layer, the single-channel grayscale image is replicated into a three-channel grayscale image. In the input-modality enhancement experiments, image gradients, edge maps, and wavelet decomposition features are introduced as additional input information. Wavelet decomposition is applied separately to the three color channels, producing four sub-band feature maps per channel. Therefore, the wavelet-feature-enhanced input contains $3+3\times4=15$ channels. These feature maps are resized to the original resolution using bilinear interpolation and concatenated with the original image along the channel dimension. Based on these color information reduction and explicit feature injection settings, this study examines whether input information changes affect model performance.

\begin{equation}
Y = 0.299R + 0.587G + 0.114B.
\label{eq:rgb2gray}
\end{equation}

\subsection{Training Settings and Evaluation Metrics}

To objectively evaluate model performance, this study uses cross-entropy loss as the optimization objective and uses test Top-1 accuracy as the primary metric for generalization performance. The detailed training configuration is shown in Table~\ref{tab:train_config}.

\begin{table}[htbp]
\centering
\caption{Training configuration for the main experiments.}
\label{tab:train_config}
\small
\begin{tabular}{ll}
\toprule
Configuration & Setting \\
\midrule
Optimizer & Adam \\
Loss function & Cross-entropy loss \\
Initial learning rate & $10^{-3}$ \\
Batch size & 128 \\
Number of epochs & 100 \\
Input size & $32 \times 32$ \\
Input normalization & mean=0.5, std=0.5 \\
Random seed & 42 \\
\bottomrule
\end{tabular}
\end{table}

During training, training loss, training accuracy, test loss, and test Top-1 accuracy are recorded. Training loss and training accuracy measure the model's fitting ability on the training data, while test loss and test accuracy evaluate generalization performance on unseen samples. This study uses test accuracy as the primary generalization metric, and test loss as an auxiliary metric to analyze changes in prediction confidence and overfitting-related phenomena.

\begin{table*}[!t]
\centering
\begin{threeparttable}
\caption{Test loss and Top-1 accuracy at the epoch with the best test accuracy under different training set sizes.}
\label{tab:size_model_test_metrics}
\scriptsize
\setlength{\tabcolsep}{4pt}
\begin{tabular}{lcc|cc|cc|cc|cc|cc}
\toprule
\multirow{2}{*}{Model} & \multirow{2}{*}{Params (M)} & \multirow{2}{*}{FLOPs (G)}
& \multicolumn{2}{c|}{5k}
& \multicolumn{2}{c|}{10k}
& \multicolumn{2}{c|}{20k}
& \multicolumn{2}{c|}{30k}
& \multicolumn{2}{c}{50k} \\
\cmidrule(lr){4-5}
\cmidrule(lr){6-7}
\cmidrule(lr){8-9}
\cmidrule(lr){10-11}
\cmidrule(lr){12-13}
& & & Loss & Acc & Loss & Acc & Loss & Acc & Loss & Acc & Loss & Acc \\
\midrule
MLP & 2.1038 & 0.0021
& 3.9727 & 0.4325
& 6.5874 & 0.4750
& 2.4204 & 0.5064
& 1.5105 & 0.5204
& 1.4065 & 0.5470 \\
AlexNet & 35.8552 & 0.2005
& 4.1721 & 0.5570
& 2.7562 & 0.6228
& 2.6117 & 0.6805
& 2.1627 & 0.7205
& 1.0352 & 0.7654 \\
ResNet-18 & 11.1740 & 0.5579
& \textbf{1.5756} & \textbf{0.6396}
& \textbf{1.3661} & \textbf{0.7384}
& 1.4130 & 0.7946
& 1.1560 & 0.8243
& 0.9969 & 0.8560 \\
ResNet-34 & 21.2821 & 1.1635
& 1.8086 & 0.6289
& 1.6392 & 0.7270
& \textbf{1.1485} & \textbf{0.8059}
& 1.0970 & 0.8309
& 0.7648 & 0.8681 \\
ResNet-50 & 23.5208 & 1.3116
& 1.9280 & 0.6094
& 1.7687 & 0.7107
& 1.3201 & 0.7924
& 1.2631 & 0.8252
& \textbf{0.8499} & \textbf{0.8709} \\
ResNet-101 & 42.5130 & 2.5304
& 1.9325 & 0.6190
& 1.6671 & 0.7202
& 1.2853 & 0.8001
& 1.1766 & 0.8267
& 0.9446 & 0.8596 \\
ResNet-152 & 58.1566 & 3.7508
& 2.1623 & 0.5907
& 1.6284 & 0.7108
& 1.2478 & 0.7906
& \textbf{0.9787} & \textbf{0.8349}
& 0.8902 & 0.8635 \\
\bottomrule
\end{tabular}

\begin{tablenotes}
\footnotesize
\item Note: Bold values indicate the best Top-1 accuracy under each training set size, together with the corresponding test loss. For each training size, Loss denotes the test loss at the epoch where the best test Top-1 accuracy is achieved.
\end{tablenotes}
\end{threeparttable}
\end{table*}

\begin{figure*}[!t]
    \centering
    \begin{subfigure}[t]{0.48\textwidth}
        \centering
        \includegraphics[width=\linewidth]{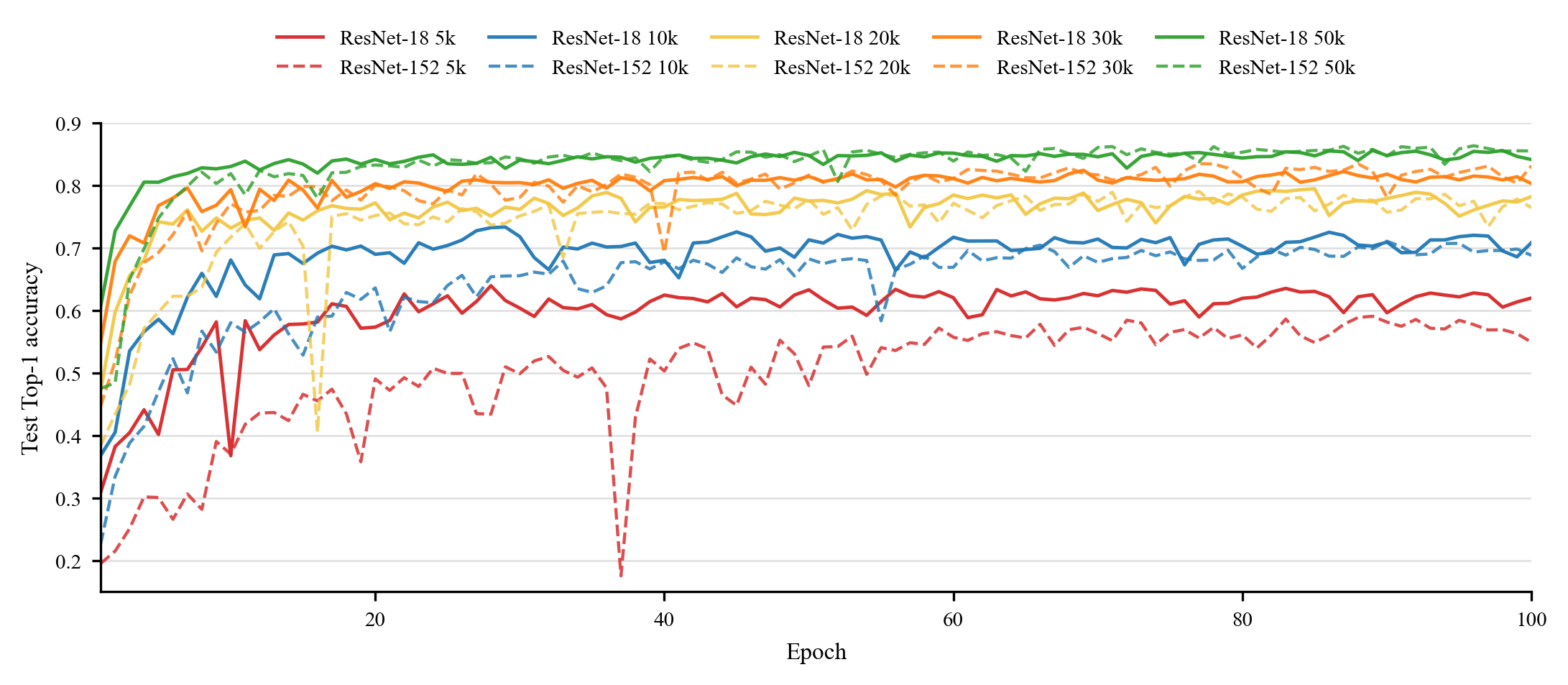}
        \caption{Test Top-1 accuracy curves of ResNet-18 and ResNet-152 under different training set sizes.}
        \label{fig:size_model_acc_curve}
    \end{subfigure}
    \hfill
    \begin{subfigure}[t]{0.48\textwidth}
        \centering
        \includegraphics[width=\linewidth]{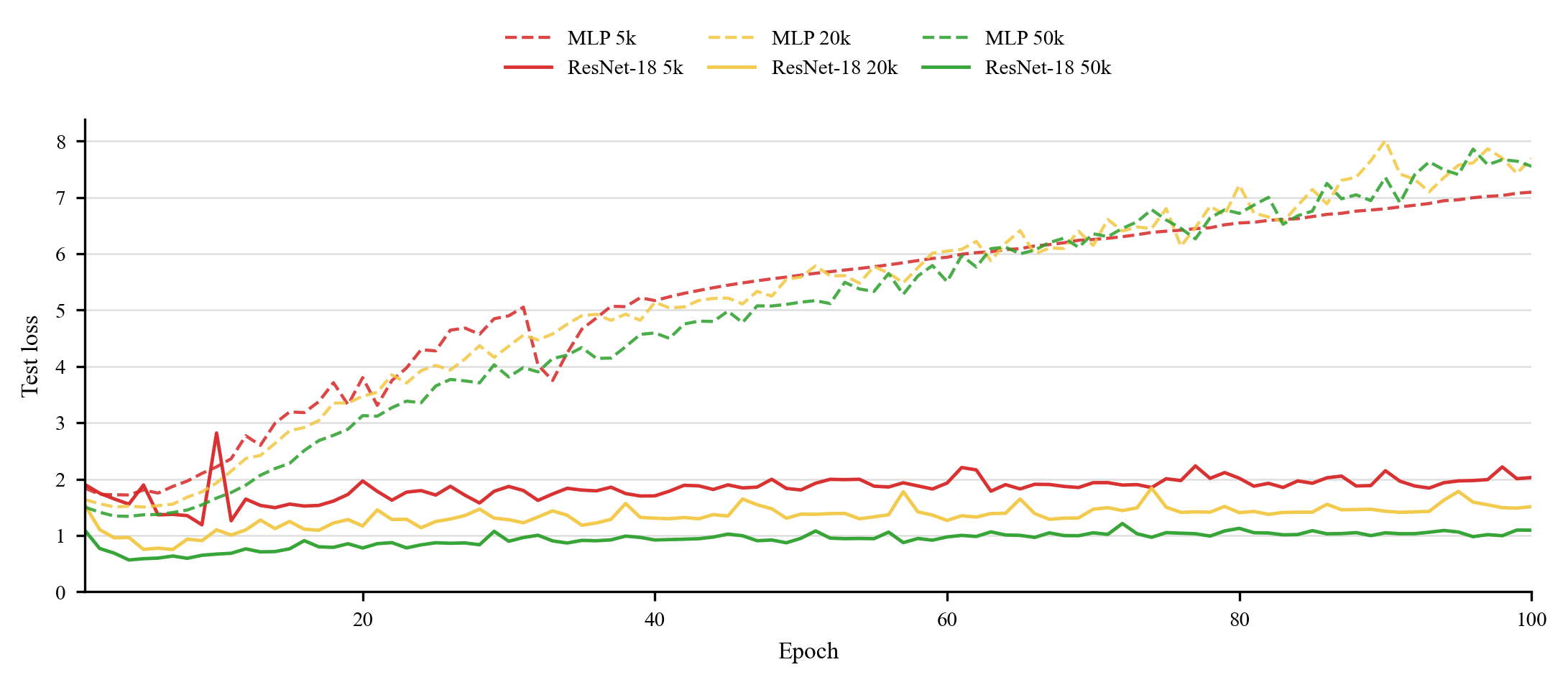}
        \caption{Test loss curves of MLP and ResNet-18 under representative training set sizes.}
        \label{fig:loss_curves}
    \end{subfigure}
    \caption{Representative test accuracy and test loss curves on CIFAR-10 across training epochs.}
    \label{fig:cifar10_curves}
\end{figure*}

% --- Section 5: Experimental Results and Quantitative Discussion ---
\section{Main Experimental Results and Analysis}

\subsection{Results and Analysis of Controlled Experiments on Data Scale and Model Complexity}

This section analyzes the joint effects of training data scale and model complexity on model generalization performance. According to the experimental design in Section~4, MLP, AlexNet, and ResNet models are trained under different training data scales, and the test Top-1 accuracy and test loss are recorded for each experimental setting. Table~\ref{tab:size_model_test_metrics} reports the best test Top-1 accuracy under different training data scales and model architectures. Since no separate validation set is used in the main experiments, the table reports the highest test accuracy and its corresponding test loss. This metric is mainly used for relative comparison across different experimental settings.

\begin{table*}[!t]
\centering
\begin{threeparttable}
\caption{Test loss and Top-1 accuracy on CIFAR-100 at the epoch with the best test accuracy under different training set sizes.}
\label{tab:cifar100_size_model_test_metrics}
\scriptsize
\setlength{\tabcolsep}{4pt}
\begin{tabular}{lcc|cc|cc|cc|cc|cc}
\toprule
\multirow{2}{*}{Model} & \multirow{2}{*}{Params (M)} & \multirow{2}{*}{FLOPs (G)}
& \multicolumn{2}{c|}{5k}
& \multicolumn{2}{c|}{10k}
& \multicolumn{2}{c|}{20k}
& \multicolumn{2}{c|}{30k}
& \multicolumn{2}{c}{50k} \\
\cmidrule(lr){4-5}
\cmidrule(lr){6-7}
\cmidrule(lr){8-9}
\cmidrule(lr){10-11}
\cmidrule(lr){12-13}
& & & Loss & Acc & Loss & Acc & Loss & Acc & Loss & Acc & Loss & Acc \\
\midrule
MLP & 2.1500 & 0.0021
& 9.7786 & 0.1605
& 10.5111 & 0.1905
& 3.5663 & 0.2175
& 3.3469 & 0.2395
& 3.2889 & 0.2645 \\
AlexNet & 36.2239 & 0.2009
& 13.8161 & 0.1693
& 11.0546 & 0.2047
& 6.4552 & 0.2980
& 4.0105 & 0.3452
& 2.2921 & 0.4324 \\
ResNet-18 & 11.2201 & 0.5579
& \textbf{3.5235} & \textbf{0.2810}
& \textbf{3.6096} & \textbf{0.3449}
& \textbf{2.7450} & \textbf{0.4598}
& \textbf{2.3198} & \textbf{0.5437}
& \textbf{2.2542} & \textbf{0.6003} \\
ResNet-34 & 21.3283 & 1.1635
& 4.2923 & 0.2682
& 4.5034 & 0.3279
& 3.1370 & 0.4383
& 3.0533 & 0.5040
& 3.4087 & 0.5748 \\
ResNet-50 & 23.7053 & 1.3117
& 8.0754 & 0.1942
& 5.3104 & 0.3130
& 4.2865 & 0.4319
& 3.9654 & 0.5025
& 3.1582 & 0.5862 \\
ResNet-101 & 42.6974 & 2.5306
& 7.6513 & 0.2055
& 4.2980 & 0.2912
& 4.1712 & 0.4262
& 3.9543 & 0.5025
& 2.8281 & 0.5840 \\
ResNet-152 & 58.3410 & 3.7510
& 7.7855 & 0.1689
& 4.8352 & 0.3032
& 4.2075 & 0.4231
& 4.1063 & 0.5124
& 3.0009 & 0.5791 \\
\bottomrule
\end{tabular}

\begin{tablenotes}
\footnotesize
\item Note: Bold values indicate the best Top-1 accuracy under each training set size, together with the corresponding test loss. For each training size, Loss denotes the test loss at the epoch where the best test Top-1 accuracy is achieved.
\end{tablenotes}
\end{threeparttable}
\end{table*}

As shown in Table~\ref{tab:size_model_test_metrics}, the test accuracy of all models improves substantially as the number of training samples increases. This indicates that, compared with increasing the parameter scale, enlarging the training data scale provides a more stable generalization gain. One possible explanation is that a larger training data scale provides stronger constraints on the model's decision function, thereby reducing the risk that the model relies on a small amount of data for memorization-based fitting. This explanation is similar to the observation in the preliminary experiment that high-degree models are more likely to exhibit severe local oscillations when sample gaps are large. However, because deep learning models have more complex internal structures, the two phenomena should not be treated as directly equivalent. Based on Table~\ref{tab:size_model_test_metrics} and the training curves, the following observations can be made.

First, increasing the number of training samples leads to a substantial improvement in generalization performance, and the test accuracy curves under different training data scales show clear stratification, as shown in Figure~\ref{fig:size_model_acc_curve}. This suggests that, under the experimental setting of this study, increasing the training data scale alone provides a more stable improvement in model generalization performance than increasing model scale alone. This result is consistent with previous studies on the complex relationship among data scale, model scale, and generalization performance~\cite{nakkiran2020deep}.

Second, even a simple MLP with three hidden layers can achieve more than 97.5\% training accuracy on the 50000-sample training set. This indicates that even a simple neural network has a sufficiently large parameter space to fit the training data well. However, its test loss and test accuracy are significantly worse than those of convolutional neural networks, suggesting that fitting ability is not necessarily strongly correlated with generalization performance. This observation is consistent with the findings of Zhang et al. on the strong fitting ability of deep networks~\cite{zhang2017understanding}.

Third, as shown in Figure~\ref{fig:loss_curves}, the MLP exhibits a continuous and substantial increase in test loss after its test accuracy becomes relatively stable. A similar phenomenon is also observed in AlexNet under the $N=5000$ setting. Since this study uses cross-entropy loss, this phenomenon may indicate that the model becomes increasingly confident on some incorrectly classified test samples, which leads to higher test loss~\cite{guo2017calibration}. Previous studies have suggested that deep networks tend to learn simple patterns first and may later gradually memorize noise or atypical patterns in the training data~\cite{arpit2017closer,rahaman2019spectral}. In contrast, the increase in test loss is milder for ResNet models. However, because ResNet introduces residual connections, batch normalization, and deeper convolutional structures, this study cannot determine whether the phenomenon is mainly caused by residual connections or by the inductive bias of convolutional computation. Further ablation studies are needed to verify this issue.

Fourth, after the number of parameters reaches a certain scale, model parameter scale does not show a stable positive correlation with improved generalization performance. As shown in Table~\ref{tab:size_model_test_metrics}, ResNet-152, which has the largest number of parameters, does not achieve the best performance in most settings. In small-sample settings, models with fewer parameters often achieve slightly better generalization performance than larger models. It is also worth noting that the stratification shown in Figure~\ref{fig:size_model_acc_curve} remains visible even between ResNet-18 and ResNet-152, despite their large difference in parameter scale. This phenomenon is also consistent with the view that model scale, data scale, and training dynamics jointly affect generalization performance~\cite{nakkiran2020deep}.

Next, to examine whether the above observations remain consistent under a more difficult task, this study applies the same experimental strategy to CIFAR-100 and further analyzes the effects of data scale and model complexity on generalization performance. CIFAR-100 contains more categories, and each category has fewer available training samples. Therefore, it is a more difficult classification task than CIFAR-10. Table~\ref{tab:cifar100_size_model_test_metrics} reports the test loss and test Top-1 accuracy of each model under different training data scales.

From Table~\ref{tab:cifar100_size_model_test_metrics} and the recorded training curves, the following observations can be made. First, on CIFAR-100, the test Top-1 accuracy of each model still improves overall as the number of training samples increases. This shows that, even under the more difficult CIFAR-100 task, increasing the training data scale still provides a relatively stable generalization gain. Meanwhile, the overall test accuracy on CIFAR-100 is clearly lower than that on CIFAR-10, indicating that higher task difficulty and fewer training samples per class reduce model generalization performance.

Second, the MLP still reaches a maximum training accuracy of 95.5\% on the full training set. This indicates that even on CIFAR-100, which has more categories, a fully connected network can still achieve high accuracy on the training set. However, the test accuracy of the MLP is significantly lower than that of convolutional neural networks in both experiments, suggesting that its fitting ability does not translate into effective generalization performance and that it exhibits a clear tendency toward overfitting.

Third, the continuous increase in test loss is not limited to the MLP. AlexNet also shows a substantial and continuous increase in test loss under all training data scales. For ResNet models, similar behavior begins to appear from ResNet-50 under the $N=5000$ setting, whereas all ResNet models maintain better test loss control under the $N=10000$ setting. This study conjectures that residual connections allow the model to learn identity mappings, enabling larger models to adaptively preserve stable training behavior and avoid degradation in deep networks. However, there may be a trade-off between overfitting and identity mapping learning. When the training data scale is small and the model scale is large, the reduced number of available training samples per class may encourage the model to further reduce training loss by exploiting sample-specific patterns or noise features, rather than learning representations that are more beneficial for generalization.

\begin{figure}[htbp]
    \centering
    \includegraphics[width=\linewidth]{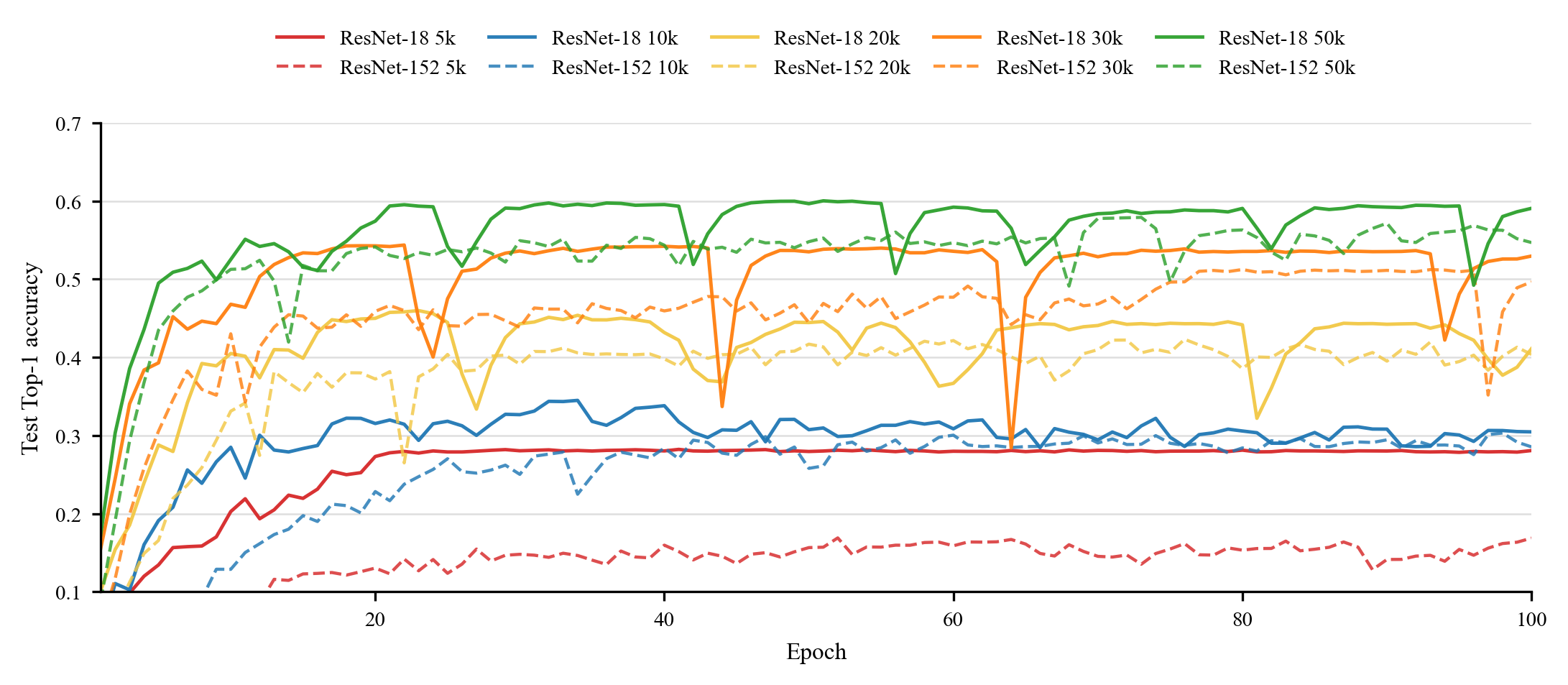}
    \caption{Test Top-1 accuracy curves of ResNet-18 and ResNet-152 under different training set sizes on CIFAR-100.}
    \label{fig:cifar100_resnet_acc_curve}
\end{figure}

Fourth, as shown in Figure~\ref{fig:cifar100_resnet_acc_curve}, although the stratification of test accuracy curves is more irregular than in CIFAR-10, it remains visible overall. The best test accuracy of ResNet-18 is higher than that of ResNet-152 in all experimental settings. By examining the experimental logs, we also observe that ResNet-152 shows a clear increase in test loss under the smallest training data scale. This study conjectures that, in settings such as CIFAR-100 with $N=5000$, where the images are small and the number of training samples is limited, the number of parameters may have a stronger effect on generalization performance. This phenomenon suggests that, for small images with limited training samples and many categories, excessive model capacity may more easily lead to overfitting. In contrast, on larger-scale datasets such as ImageNet, the training data provide richer visual variations and category information~\cite{deng2009imagenet}, making it easier for deep models to learn general-purpose feature representations. Therefore, whether model complexity provides a generalization gain likely depends on the alignment among data scale, task difficulty, and model architecture.

\subsection{Results and Analysis of Controlled Experiments on Input Modalities}

\begin{table*}[!t]
\centering
\begin{threeparttable}
\caption{Performance comparison under different input information settings on CIFAR-10 and CIFAR-100.}
\label{tab:input_mode_metrics_both}
\small
\setlength{\tabcolsep}{5pt}
\begin{tabular}{llc|cc|cc}
\toprule
\multirow{2}{*}{Model} & \multirow{2}{*}{Input mode} & \multirow{2}{*}{Channels}
& \multicolumn{2}{c|}{CIFAR-10}
& \multicolumn{2}{c}{CIFAR-100} \\
\cmidrule(lr){4-5}
\cmidrule(lr){6-7}
& & & Loss & Acc & Loss & Acc \\
\midrule
MLP & RGB & 3 & 1.6158 & 0.5518 & 3.2889 & 0.2645 \\
MLP & Gray & 3 & 2.0550 & 0.4591 & 3.9983 & 0.1789 \\
MLP & RGB+Gradient & 4 & 4.9993 & 0.5741 & 3.1528 & 0.2933 \\
MLP & RGB+Edge & 4 & 1.4887 & 0.5617 & 3.2992 & 0.2835 \\
MLP & RGB+Wavelet & 15 & \textbf{1.2822} & \textbf{0.6119} & \textbf{3.1606} & \textbf{0.3107} \\
\midrule
ResNet-18 & RGB & 3 & \textbf{0.8830} & \textbf{0.8677} & \textbf{2.2542} & \textbf{0.6003} \\
ResNet-18 & Gray & 3 & 1.0849 & 0.8432 & 2.6995 & 0.5527 \\
ResNet-18 & RGB+Gradient & 4 & 0.9529 & 0.8655 & 2.2802 & 0.5924 \\
ResNet-18 & RGB+Edge & 4 & 0.9646 & 0.8632 & 2.2162 & 0.5956 \\
ResNet-18 & RGB+Wavelet & 15 & 0.9621 & 0.8501 & 2.3427 & 0.5786 \\
\midrule
ResNet-50 & RGB & 3 & 0.8185 & 0.8707 & \textbf{3.1582} & \textbf{0.5862} \\
ResNet-50 & Gray & 3 & 1.1017 & 0.8467 & 4.0404 & 0.5286 \\
ResNet-50 & RGB+Gradient & 4 & 0.8460 & 0.8727 & 3.2978 & 0.5704 \\
ResNet-50 & RGB+Edge & 4 & 0.9701 & 0.8603 & 3.3087 & 0.5677 \\
ResNet-50 & RGB+Wavelet & 15 & \textbf{0.8399} & \textbf{0.8731} & 3.4775 & 0.5612 \\
\bottomrule
\end{tabular}

\begin{tablenotes}
\footnotesize
\item Note: Bold values indicate the best Top-1 accuracy for each model on each dataset, together with the corresponding test loss. Gray denotes the three-channel grayscale input. RGB+Gradient, RGB+Edge, and RGB+Wavelet denote RGB inputs concatenated with gradient, edge, and wavelet feature maps, respectively. Loss denotes the test loss at the epoch where the best test Top-1 accuracy is achieved.
\end{tablenotes}
\end{threeparttable}
\end{table*}

This section further analyzes the effect of changes in input modalities on model performance. According to the experimental method described in Section~4, RGB image input is used as the baseline, and grayscale input, gradient-enhanced input, edge-enhanced input, and wavelet-feature-enhanced input are constructed separately~\cite{canny1986computational,mallat1989theory}. These settings are used to examine whether color information reduction and the introduction of explicit prior features affect model fitting ability and generalization performance. In terms of model selection, based on the preceding experimental results, ResNet-18 has already shown strong generalization performance on both datasets. Therefore, this study selects MLP, ResNet-18, and ResNet-50 as representative models to compare how different model architectures and levels of model complexity respond to changes in input modalities.

As shown in Table~\ref{tab:input_mode_metrics_both}, grayscale input leads to a decrease in test accuracy across all models and datasets. This indicates that color information remains an effective feature for CIFAR image classification tasks and can be utilized by both MLP and convolutional neural networks. In contrast, explicit prior features such as gradients, edges, and wavelets provide relatively clear improvements for MLP, but do not bring stable gains for ResNet models; in some settings, they even lead to performance degradation. This suggests that, under the experimental setting of this study, the effect of explicit prior features is closely related to model architecture. For fully connected networks that lack convolutional inductive bias, additional prior features can provide useful auxiliary information. For convolutional networks, however, the model can already learn local textures, edges, and shape-related information from the original images, so simple channel concatenation does not necessarily lead to stable improvements.

\subsection{Experimental Summary and Discussion}

The experimental results show that training data scale, model complexity, and input modalities do not affect generalization performance independently. Increasing the number of training samples usually provides more stable generalization gains; the benefit of model complexity depends on data scale and task difficulty; and whether changes in input modalities are effective also depends on whether the model architecture can utilize the additional information. For MLP, which lacks image-specific inductive bias, additional prior features can provide useful auxiliary information. For convolutional neural networks, however, simple channel-concatenation-based prior injection does not provide stable improvements. This result suggests that improving generalization performance depends not only on increasing the number of training samples, the number of parameters, or the amount of input information, but also on the alignment between model architecture and data characteristics.

Although additional input modalities do not provide stable improvements for convolutional models in this study, this does not imply that explicit prior information has no value. A possible reason is that this study introduces prior information through direct channel concatenation, which may not be well aligned with the feature extraction mechanism of convolutional neural networks. For convolutional neural networks that already possess image-specific inductive biases such as local connectivity and weight sharing~\cite{goodfellow2016deep,geirhos2019imagenet}, simply concatenating gradient, edge, or wavelet features may not allow the model to use these features effectively. Therefore, improving model generalization performance requires not only increasing training data scale, but also designing model architectures that are better aligned with the structure of the data and the form of prior information. Existing studies have shown that, in some tasks, introducing prior knowledge through attention mechanisms, multimodal fusion, or physical constraints can improve learning ability and model performance~\cite{hu2018squeeze,baltrusaitis2019multimodal,raissi2019physics}. Compared with simple channel concatenation, these methods usually incorporate prior information into the model architecture or loss function, thereby forming a more effective inductive bias.

% --- Section 6: Conclusion ---
\section{Conclusion}

This study investigates the effects of training data scale, model complexity, and input modalities on model performance, especially generalization performance, through a preliminary experiment and a series of main experiments. The experimental results show that increasing the training data scale consistently improves test accuracy on both CIFAR-10 and CIFAR-100, indicating that training data scale remains an important factor affecting model generalization performance. In contrast, increasing model complexity does not consistently lead to improved generalization performance. The input-modality experiments further suggest that a model's inductive bias may play a key role in generalization performance. In other words, whether a model can effectively utilize input information is an important factor influencing its generalization performance. Overall, improving generalization performance depends not only on increasing the number of training samples, the number of parameters, or the amount of input information, but also on the alignment between model architecture and data characteristics.

Nevertheless, this study has several limitations. First, the experiments are mainly conducted on CIFAR-10 and CIFAR-100, two small-resolution image classification datasets. Although these datasets allow comparison under different levels of task difficulty, the conclusions still need to be further validated on larger-scale and more complex datasets. Second, to focus on the effects of data scale, model complexity, and input modalities themselves, this study does not use common training strategies such as data augmentation, weight decay, Dropout, or learning-rate scheduling. Therefore, the experimental results mainly reflect phenomena under a controlled experimental setting. In addition, the experiments are mainly conducted with a single random seed, and mean and variance over multiple repeated runs are not reported. Finally, this study introduces explicit prior features mainly through channel concatenation, and does not further investigate more complex prior-injection methods such as attention-based fusion, structured prior constraints, or frequency-domain neural networks. Future work can further analyze the interactions among data scale, model complexity, and input information structure using larger-scale datasets, richer model architectures, and more systematic prior-fusion methods, thereby providing a deeper understanding of the sources of deep neural network generalization performance.

\bibliographystyle{unsrt}
\bibliography{references}

\end{document}